\documentclass[runningheads]{llncs}
\usepackage{bm}
\usepackage{booktabs}
\usepackage{mathtools}
\usepackage{amsfonts}
\usepackage{dsfont}
\usepackage[flushleft]{threeparttable}
\usepackage{xcolor}
\usepackage{graphicx}
\usepackage{caption}
\usepackage{subcaption}
\usepackage{hyperref}
\usepackage{cleveref}

\newcommand{\hide}[1]{}

\usepackage{xcolor}
\usepackage{amssymb}

\definecolor{lennartlmucolor}{rgb}{0,0.7,0.3}
\definecolor{lennarttudcolor}{rgb}{0,0.3,0.7}
\definecolor{raphaelcolor}{rgb}{0.6,0.8,0.2}
\definecolor{pascalcolor}{rgb}{0.2,0.6,0.6}
\definecolor{heikecolor}{rgb}{1,0.75,0.8}
\definecolor{berndcolor}{rgb}{1, 0.65, 0}
\definecolor{todocolor}{rgb}{0.9,0.1,0.1}
\definecolor{changedcolor}{rgb}{0.42,0.27,0.57}
\definecolor{addedcolor}{rgb}{0.867,0.176,0.361}

\newcommand{\redacted}[1]{\emph{[anonymized for review]}}

\ifdefined\N                                                                
\renewcommand{\N}{\mathds{N}}                                                
\else
  \newcommand{\N}{\mathds{N}}
\fi
\ifdefined\C
  \renewcommand{\C}{\mathds{C}}                                             
\else
  \newcommand{\C}{\mathds{C}}
\fi

\DeclareMathOperator*{\argmin}{arg\,min}

\newcommand{\allDatasets}{\mathds{D}}                                       
\newcommand{\D}{\mathcal{D}}                                                      
\renewcommand{\xi}[1][i]{\mathbf{x}^{(#1)}}                                          

\newcommand{\preimageInducerShort}{\allDatasets\times\Lambda}     
\newcommand{\inducer}{\mathcal{I}}                                                

\newcommand{\Hspace}{\mathcal{H}}														
\newcommand{\fh}{\hat{f}}                                                   
\newcommand{\GEh}{\widehat{\mathrm{GE}}}                                             
           
\newcommand{\GEhlam}{\GEh(\lambdav)}                                                 

\newcommand{\lambdav}{\bm{\lambda}}											

\newcommand{\Ilam}{\inducer_{\lambdav}}						
\newcommand{\lams}{\lambdav^{*}}		                    
\newcommand{\LamS}{\tilde\Lambda}                           

\newcommand{\online}{\href{https://github.com/slds-lmu/hpo_ela}{online appendix}}

\begin{document}
\title{\texorpdfstring{HPO $\times$ ELA: Investigating Hyperparameter Optimization Landscapes by Means of Exploratory Landscape Analysis}{HPO X ELA: Investigating Hyperparameter Optimization Landscapes by Means of Exploratory Landscape Analysis}}
\titlerunning{HPO $\times$ ELA}
%
\author{Lennart Schneider\inst{1}\thanks{Equal contributions.}\orcidID{0000-0003-4152-5308} \and
Lennart Schäpermeier\inst{2}$^\star$\orcidID{0000-0003-3929-7465} \and
Raphael Patrick Prager\inst{3}$^\star$\orcidID{0000-0003-1237-4248} \and
Bernd Bischl\inst{1}\orcidID{0000-0001-6002-6980} \and
Heike Trautmann\inst{3,4}\orcidID{0000-0002-9788-8282} \and
Pascal Kerschke\inst{2}\orcidID{ 0000-0003-2862-1418}
}
\authorrunning{L. Schneider et al.}
%
\institute{
Chair of Statistical Learning and Data Science,
LMU Munich,
Germany \\
\texttt{\email{\{lennart.schneider,bernd.bischl\}@stat.uni-muenchen.de}}
\and
Big Data Analytics in Transportation,
TU Dresden,
Germany \\
\texttt{\email{\{lennart.schaepermeier,pascal.kerschke\}@tu-dresden.de}}
\and
Data Science: Statistics and Optimization,
University of M{\"u}nster,
Germany \\
\texttt{\email{\{raphael.prager,heike.trautmann\}@wi.uni-muenster.de}}
\and
Data Management and Biometrics Group,
University of Twente, Netherlands
}
\maketitle              
\begin{abstract}
Hyperparameter optimization (HPO) is a key component of machine learning models for achieving peak predictive performance.
While numerous methods and algorithms for HPO have been proposed over the last years, little progress has been made in illuminating and examining the actual structure of these black-box optimization problems.
Exploratory landscape analysis (ELA) subsumes a set of techniques that can be used to gain knowledge about properties of unknown optimization problems.
In this paper, we evaluate the performance of five different black-box optimizers on $30$ HPO problems, which consist of two-, three- and five-dimensional continuous search spaces of the XGBoost learner trained on $10$ different data sets.
This is contrasted with the performance of the same optimizers evaluated on $360$ problem instances from the black-box optimization benchmark (BBOB).
We then compute ELA features on the HPO and BBOB problems and examine similarities and differences.
A cluster analysis of the HPO and BBOB problems in ELA feature space allows us to identify how the HPO problems compare to the BBOB problems on a structural meta-level.
We identify a subset of BBOB problems that are close to the HPO problems in ELA feature space and show that optimizer performance is comparably similar on these two sets of benchmark problems.
We highlight open challenges of ELA for HPO and discuss potential directions of future research and applications.

\end{abstract}

\section{Introduction}

In machine learning (ML), hyperparameter optimization (HPO) constitutes one of the most frequently used tools for improving the predictive performance of a model \cite{bischl_hyperparameter_2021}.
The goal of classical single-objective HPO is to find a hyperparameter configuration that minimizes the estimated generalization error.
Generally, neither a closed-form mathematical representation nor analytic gradient information is available, making HPO a \emph{black-box} optimization problem and evolutionary algorithms (EAs) and model-based optimizers good candidate algorithms. 
As a consequence, no prior information about the optimization landscape -- which could allow comparisons of HPO and other black-box problems, or provide guidance regarding the choice of optimizer -- is available.
This also extends to automated ML (AutoML) \cite{hutter2019automated}, which builds upon HPO. 

In contrast, in the domain of \emph{continuous} black-box optimization, a sophisticated toolbox for landscape analysis and the characterization of their properties has been developed over the years.
In exploratory landscape analysis (ELA), optimization landscape features are calculated from small samples of evaluated points from the original black-box problem.
It has been shown in numerous studies that ELA feature sets capture relevant landscape characteristics and that they can be used for automated algorithm selection, improving upon the state-of-the-art selector \cite{bischl2012algorithm,kerschke2019automated}.
Particularly well-studied are the functions from the black-box optimization benchmark (BBOB) \cite{hansen2009real}.

Empirical studies \cite{pushak2018algorithm,pushak2020golden} in the closely related area of algorithm configuration hint that performance landscapes often are rather benign, i.e., unimodal and convex, although this only holds for an aggregation over larger instance sets and their analysis does not allow further characterization of individual problem landscapes.
There exists some work to circumvent HPO altogether, by automatically configuring an algorithm for a given problem instance \cite{belkhir2017per,prager2020per}.
However, these are limited to configuring optimization algorithms rather than ML models.
In addition, they are often restricted in the number and type of variables they are able to configure. 
\cite{pimenta2020fitness} apply fitness landscape analysis on AutoML landscapes, computing fitness distance correlations and neutrality ratios on various AutoML problems.
They utilize these features only in an exploratory manner, characterizing the landscapes, without a link to optimizer performance, and cannot compare the analyzed landscapes to other black-box problems in a natural way.
Similar work on fitness landscape analysis exists but focuses mostly on neural networks \cite{Bosman2018,traore2021}.
Some preliminary work \cite{doerr2019making} on the hyperparameters of a $(1+1)$-EA on a OneMax problem suggests that the ELA feature distribution of a HPO problem can be significantly different from other benchmark problems.
Recently, \cite{pushak2022automl} developed statistical tests for the deviation of loss landscapes from uni-modality and convexity and showed that loss landscapes of AutoML problems are highly structured and often uni-modal.

In this work, we characterize continuous HPO problems using ELA features, enabling comparisons between different black-box optimization problems and optimizers. Our main contributions are as follows:
\begin{enumerate}
    \item We examine similarities and differences of HPO and BBOB problems by investigating the performance of different black-box optimizers. 
    \item We compute ELA features for all HPO and BBOB problems and demonstrate their usefulness in distinguishing between HPO and BBOB.
    \item We demonstrate how HPO problems position themselves in ELA feature space on a meta-level by performing a cluster analysis on principle components derived from ELA features of HPO and BBOB problems and investigate performance differences of optimizers on HPO problems and BBOB problems that are close to the HPO problems in ELA feature space.
    \item We discuss how ELA can be used for HPO in future work and highlight open challenges of ELA in the context of HPO.
    \item We release code and data of all our benchmark experiments hoping to facilitate future research (which currently may be hindered due to the computationally expensive HPO black-box evaluations).
\end{enumerate}
The remainder of this paper is structured as follows: Fundamentals for HPO and ELA are introduced in \Cref{sec:background}. The experimental setup is presented in \Cref{sec:setup}, with the results regarding the algorithm performance and ELA feature space analysis in \Cref{sec:performance,sec:ela_analysis}, respectively. \Cref{sec:conclusion} concludes this paper and offers future research directions.



\section{Background} \label{sec:background}

\paragraph{Hyperparameter Optimization}

Hyperparameter optimization (HPO) methods aim to identify a well-performing hyperparameter configuration $\lambdav \in \LamS$ for an ML algorithm $\Ilam$ \cite{bischl_hyperparameter_2021}.
An ML \emph{learner} or \emph{inducer} $\inducer$ configured by hyperparameters $\lambdav \in \Lambda$ maps a data set $\D \in \allDatasets$ to a model $\fh$, i.e.,
$
\inducer : \preimageInducerShort \to \Hspace, 
(\D, \lambdav) \mapsto \fh
$.
$\Hspace$ denotes the so-called hypothesis space, i.e., the function space to which a model belongs \cite{bischl_hyperparameter_2021}.
The considered search space $\LamS \subset \Lambda$ is typically a subspace of the set of all possible hyperparameter configurations: $\LamS = \LamS_1 \times \LamS_2 \times \dots \times \LamS_d,$ where $\LamS_i$ is a bounded subset of the domain of the $i$-th hyperparameter $\Lambda_i$.
This $\LamS_i$ can be either real, integer, or category valued, and the search space can contain dependent hyperparameters, leading to a possibly hierarchical search space.
The classical (single-objective) HPO problem is defined as:
\begin{eqnarray}
    \lams \in \argmin_{\lambdav \in \LamS} \GEhlam,
    \label{eq:hpo_objective}
\end{eqnarray}
i.e., the goal is to minimize the estimated generalization error.
This typically involves a costly resampling procedure that can take a significant amount of time, see \cite{bischl_hyperparameter_2021} for further details.
$\GEhlam$ is a black-box function, as it generally has no closed-form mathematical representation, and analytic gradient information is generally not available. 
Therefore, the minimization of $\GEhlam$ forms an \emph{expensive black-box} optimization problem.
In general, $\GEhlam$ is only a stochastic estimate of the true unknown generalization error.
Formally, $\GEhlam$ depends on the concrete inducer, a resampling strategy (e.g., cross-validation) and a performance metric, for more details see \cite{bischl_hyperparameter_2021}.
In the following, we use the logloss as performance metric:
\begin{eqnarray}
\frac{1}{n_{\text{test }}} \sum_{i=1}^{n_{\text{test}}}\left(-\sum_{k=1}^{g} \sigma_{k}\left(y^{(i)}\right) \log \left(\hat{\pi}_{k}\left(\mathbf{x}^{(i)}\right)\right)\right).
\end{eqnarray}
Here, $g$ is the total number of classes, $\sigma_{k}\left(y^{(i)}\right)$ is 1 if $y$ is class $k$, and $0$ otherwise (multi-class one-hot encoding), and $\hat{\pi}_{k}\left(\mathbf{x}^{(i)}\right)$ is the estimated probability for observation $\mathbf{x}^{(i)}$ belonging to class $k$.

\paragraph{Exploratory Landscape Analysis}

The optimization landscapes of black-box functions, by design, carry no prior problem information, beyond the definition of their search parameters, which can be used for their characterization.
In the continuous domain, ELA \cite{Mersmann_Bischl_Trautmann_Preuss_Weihs_Rudolph_2011} addresses this problem by computing features on a small sample of evaluated points, which can be used for better understanding optimizer performance \cite{mersmann2015analyzing}, algorithm selection \cite{kerschke2019automated} and even algorithm configuration \cite{prager2020per}.

The original ELA features consist, e.g., of meta model features (\texttt{ela\_meta}) such as adjusted $R^2$ values for quadratic and linear models and $y$-distribution features (\texttt{ela\_distr}) such as the skewness and kurtosis of the objective values.
Over time, researchers continued to propose further feature sets, including nearest better clustering (\texttt{nbc}) \cite{kerschke2015detecting} and dispersion (\texttt{disp}) \cite{lunacek2006dispersion} features to measure multi-modality, and information content (\texttt{ic}) features \cite{munoz2014exploratory}, which extract features from random walks across the problem landscape. The R package \texttt{flacco} \cite{kerschke2019comprehensive} and Python package \texttt{pflacco} \cite{prager2022pflacco} implement a collection of the most widely used ELA feature sets. 

ELA studies often focus on the noiseless BBOB functions, as they offer diverse, well-understood challenges (such as conditioning and multimodality) and a wide range of algorithm performance data is readily available.
BBOB consists of 24 minimization problems, which are identified by their function ID (FID) and scalable with respect to their dimensionality, which ranges from $2$ to $40$.
Furthermore, different instances, identified by instance IDs (IIDs), are defined for each function, creating slightly different optimization problems with the same fundamental characteristics by means of randomized transformations in the decision and objective space.
All $D$-dimensional BBOB problems share a decision space of $[-5,5]^D$, which is guaranteed to contain the (known) optimum.

\section{Experimental Setup} \label{sec:setup}

We compare the following optimizers:
\texttt{CMAES} (a simple CMA-ES with $\sigma_{0} = 0.5$ and no restarts), \texttt{GENSA} (a generalized simulated annealing approach as described in \cite{xiang2013}), \texttt{Grid} (a grid search performed by generating a uniform sized grid over the search space and evaluating configurations of the grid in random order), \texttt{Random} (random search performed by sampling configurations uniformly at random), and \texttt{MBO} (Bayesian optimization using a Gaussian process as surrogate model and expected improvement as acquisition function \cite{jones98}, similarly configured as in \cite{riche_2021}). All optimizers were given a budget of $50D$ function evaluations in total (where $D$ is the dimensionality of the problem). All optimizer runs were replicated $10$ times.
We choose these optimizers for the following reasons: (1) they cover a wide range of optimizers that can be used for a black-box problem, (2) \texttt{Grid} and especially \texttt{Random} are frequently used for HPO and \texttt{Random} often can be considered a strong baseline \cite{bergstra_algorithms_2011}.

As HPO problems, we tune XGBoost\footnote{using a \texttt{gbtree} booster} \cite{chen2016} on ten different OpenML \cite{vanschoren2013openML} data sets (classification tasks) chosen from the OpenML-CC18 benchmarking suite \cite{bischl_2021}.
The specific data sets were chosen to cover a variety of the number of classes, instances, and features (cf.~\Cref{tab:openml}).
To reduce noise as much as possible, performance (logloss) is estimated via 10-fold cross-validation with a fixed instantiating per data set.
On each data set, we create $2, 3$ and $5$ dimensional XGBoost problems by tuning \texttt{nrounds}, \texttt{eta} ($2D$), \texttt{lambda} ($3D$), \texttt{gamma} and \texttt{alpha} ($5D$), resulting in $30$ problems in total.
We selected these hyperparameters because (1) they can be incorporated in a purely continuous search space which is generally required for the computation of ELA features, (2) they have been shown to be influential on performance \cite{probst2019} and (3) have a straightforward interpretation, i.e., \texttt{nrounds} controls the number of boosting iterations (typically increasing performance but also the tendency to overfit) while the other hyperparameters counteract overfitting and control various aspects of regularization.
The full search space is described in \Cref{tab:ss_xgboost}.
Note that \texttt{nrounds} is tuned on a logarithmic scale and therefore all parameters are treated as continuous during optimization.
Missing values of numeric features were imputed using Histogram imputation (values are drawn uniformly at random between lower and upper histogram breakpoints with cells being sampled according to the relative frequency of points contained in a cell).
Missing values of factor variables were imputed by adding a new factor level and factor variables were encoded using one-hot-encoding.
While XGBoost is a practically relevant learner we do have to note that only considering a single learner is somewhat restrictive.
We discuss this limitation in \Cref{sec:conclusion}.
In the following, individual HPO problems are abbreviated by \texttt{<name>\_<d>}, i.e., \texttt{wilt\_2} for the $2D$ wilt problem.

\begin{table}[!t]
\begin{minipage}[t]{0.46\textwidth}
  \centering
  \caption{OpenML data sets.}
  \label{tab:openml}
  \footnotesize
  \begin{threeparttable}
  \begin{tabular}{llrrr}
  \toprule
  & & \multicolumn{3}{c}{\textbf{Number of}} \\
  \textbf{ID} & \textbf{Name} & \textbf{Cl.} & \textbf{Inst.} & \textbf{Feat.} \\
  \midrule
  40983     & wilt              & 2 & 4839  & 5\\
  469       & analcatdata\_dmft & 6 & 797   & 4\\
  41156     & ada               & 2 & 4147  & 48\\
  6332      & cylinder-bands    & 2 & 540   & 37\\
  23381     & dresses-sales     & 2 & 500   & 12\\
  1590      & adult             & 2 & 48842 & 14\\
  1461      & bank-marketing    & 2 & 45211 & 16\\
  40975     & car               & 4 & 1728  & 6\\
  41146     & sylvine           & 2 & 5124  & 20\\  
  40685     & shuttle           & 7 & 58000 & 9\\
  \bottomrule
  \end{tabular}
  \begin{tablenotes}
  \tiny
  \item IDs correspond to OpenML data set IDs, which enable to query data set properties via \url{https://www.openml.org/d/<id>}.
  \end{tablenotes}
  \end{threeparttable}
\end{minipage}
\hfill
\begin{minipage}[t]{0.52\textwidth}
  \centering
  \caption{XGBoost search space.}
  \label{tab:ss_xgboost}
  \footnotesize
  \begin{threeparttable}
  \begin{tabular}{lrrr}
  \toprule
  \textbf{Hyper-} & && \\
  \textbf{param.} & \textbf{Type} & \textbf{Range} & \textbf{Trafo} \\
  \midrule
  nrounds & int. & $[3, 2000]$ & log \\
  eta & cont. & $[\exp(-7), \exp(0)]$ & log \\ 
  lambda & cont. & $[\exp(-7), \exp(7)]$ & log\\
  gamma & cont. & $[\exp(-10), \exp(2)]$ & log \\
  alpha & cont. & $[\exp(-7), \exp(7)]$ & log \\
  \bottomrule
  \end{tabular}
  \begin{tablenotes}
  \tiny
  \item ``log'' in the Trafo column indicates that this parameter is optimized on a (continuous) logarithmic scale, i.e., the range is given by $[\log(\mathrm{lower}), \log(\mathrm{upper})]$, and values are re-transformed via the exponential function prior to their evaluation. Parameters part of the full XGBoost search space that are not shown are set to their default.
  \end{tablenotes}
  \end{threeparttable}
\end{minipage}
\end{table}

As BBOB problems we select FIDs $1-24$ with IIDs $1-5$ with a dimensionality of $\{2,3,5\}$, resulting in $360$ problems in total.
We abbreviate individual BBOB problems by \texttt{<fid>\_<iid>\_<dim>}, i.e., \texttt{24\_1\_5} for FID $24$ with IID $1$ in the $5D$ setting.
Experiments have been conducted in R \cite{R}, where the individual implementation of an optimizer is referenced in the \texttt{mlr3} ecosystem \cite{mlr3}.
The package \texttt{smoof} \cite{smoof} provides the aforementioned BBOB problems.
We release all data and code for running the benchmarks and analyzing results via the following GitHub repository: \url{https://github.com/slds-lmu/hpo_ela}.
HPO benchmarks took around 2.2 CPU years on Intel Xeon E5-2670 instances, with optimizer overhead ranging from $10\%$ (\texttt{MBO} for $5D$) to less than $1\%$ (\texttt{Random} or \texttt{Grid}).

\section{Optimizer Performance} \label{sec:performance}

\begin{figure}[!t]
  \centering
  \begin{subfigure}[t]{0.32\linewidth}
    \centering
    \includegraphics[width=\textwidth]{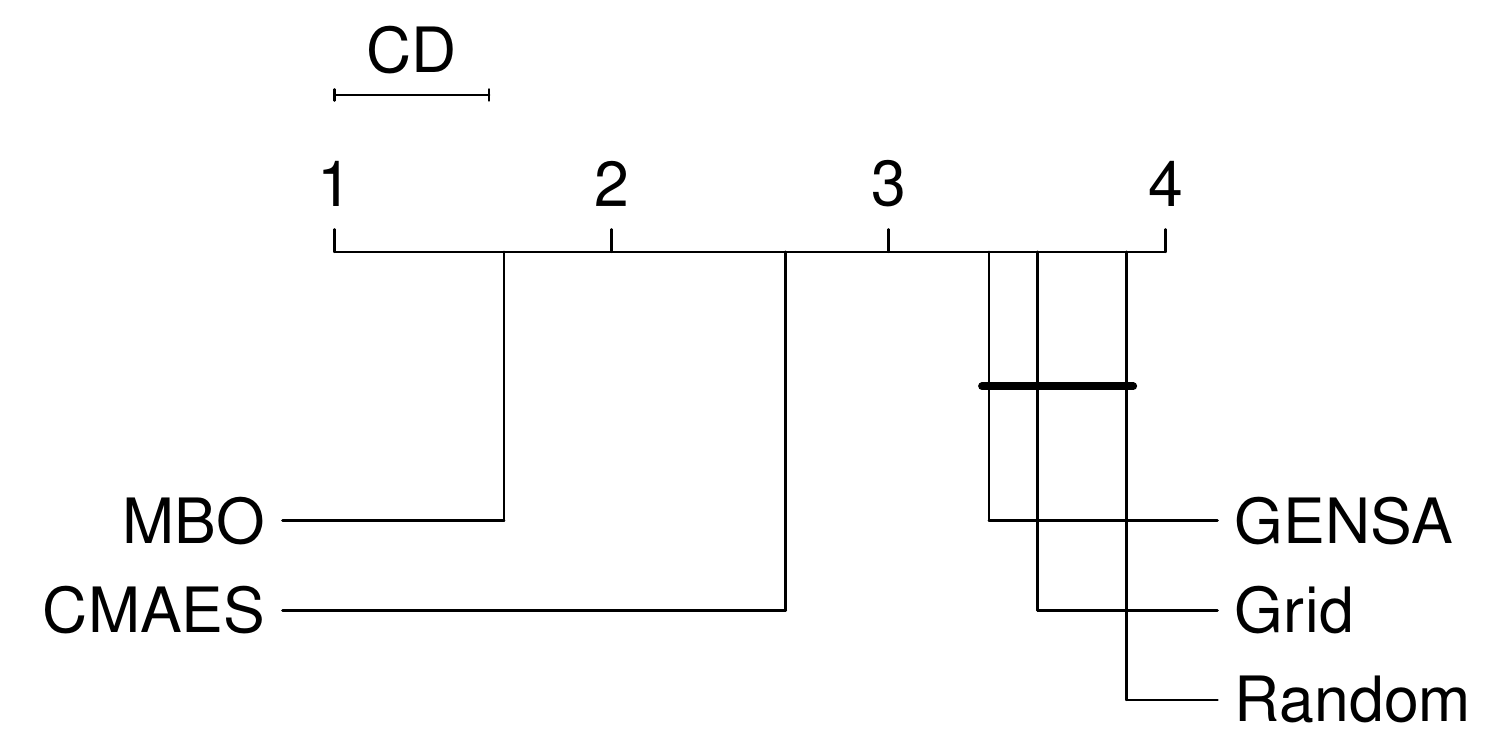}
    \caption{BBOB $2D$.}
    \label{fig:bbob_2_cd}
  \end{subfigure}
  \hfill
  \begin{subfigure}[t]{0.32\linewidth}
    \centering
    \includegraphics[width=\textwidth]{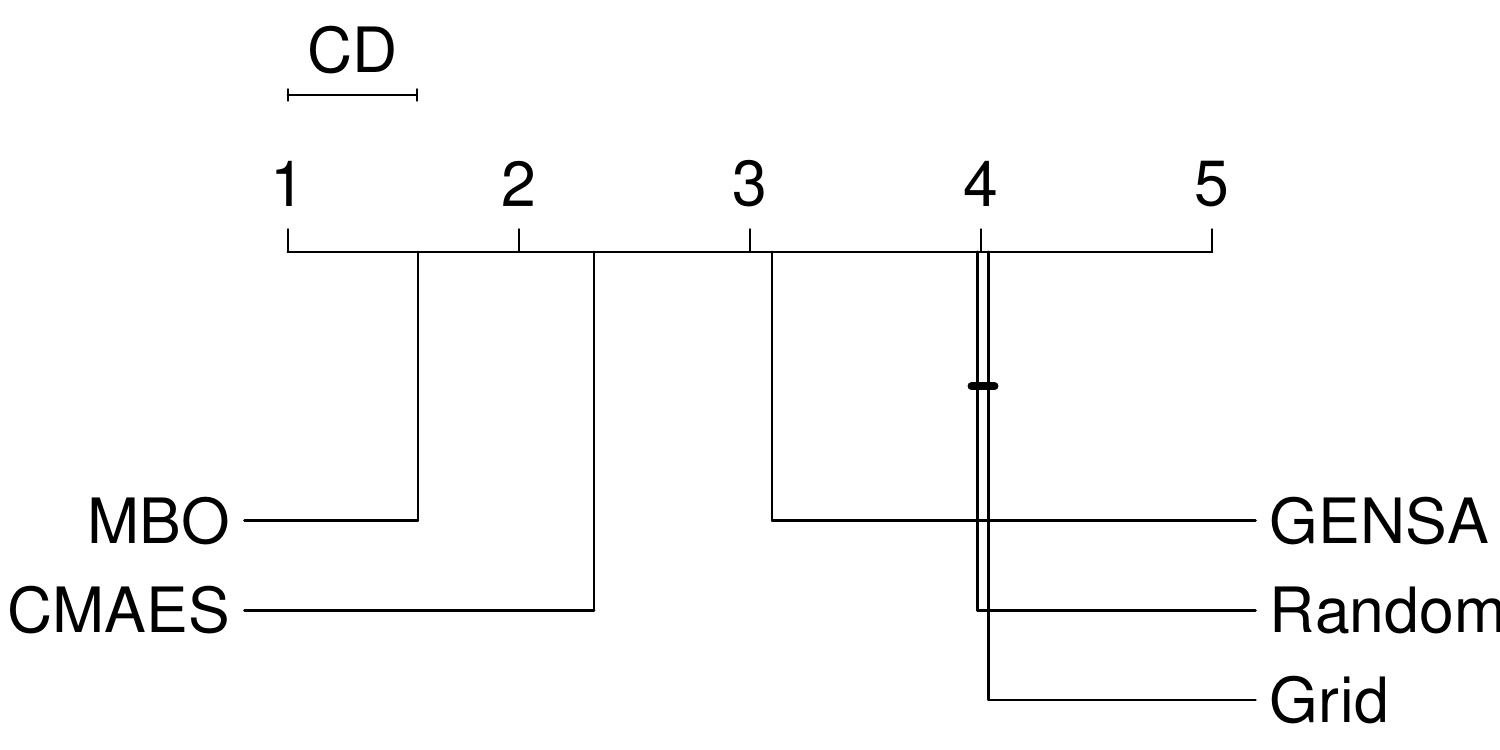}
    \caption{BBOB $3D$.}
    \label{fig:bbob_3_cd}
  \end{subfigure}
  \hfill
  \begin{subfigure}[t]{0.32\linewidth}
    \centering
    \includegraphics[width=\textwidth]{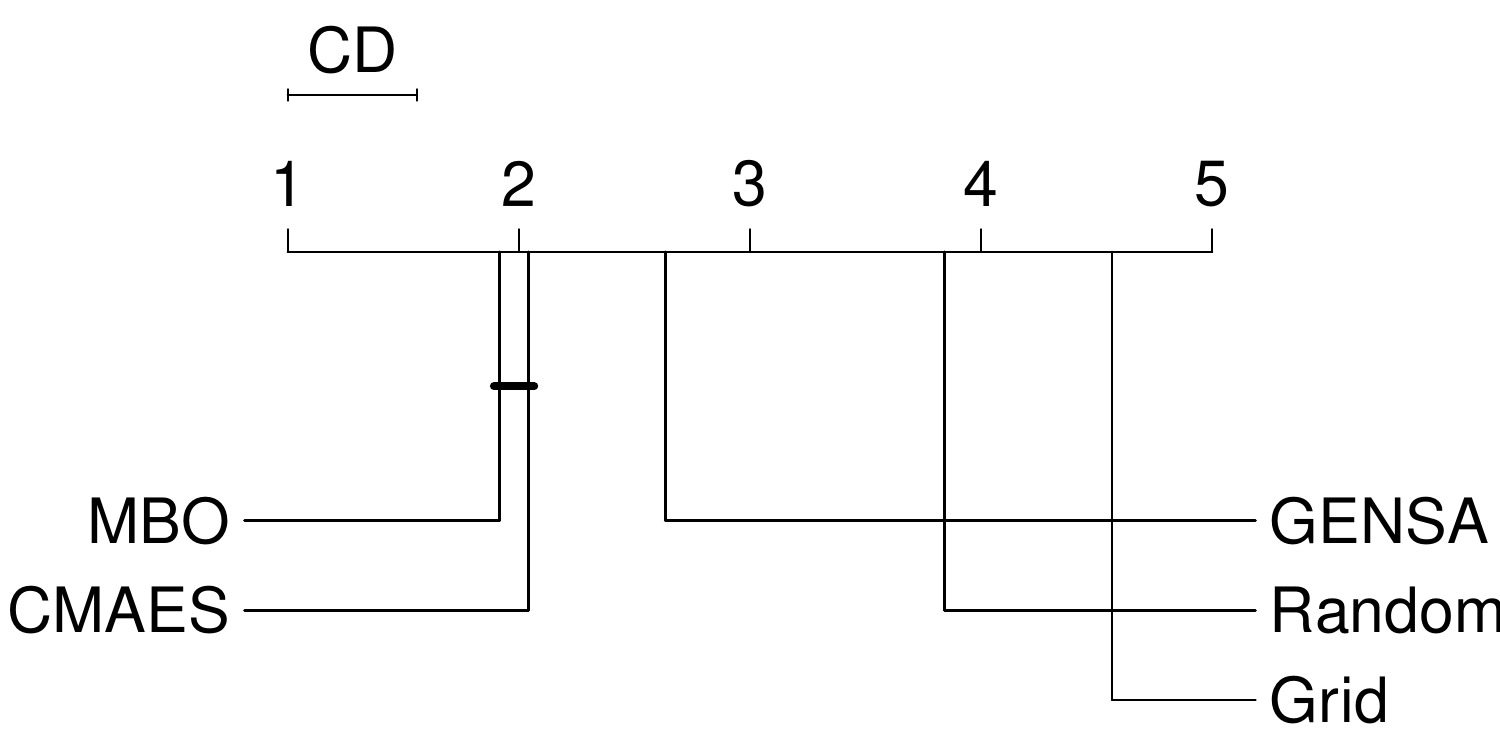}
    \caption{BBOB $5D$.}
    \label{fig:bbob_5_cd}
  \end{subfigure}
  \vfill
  \begin{subfigure}[t]{0.32\linewidth}
    \centering
    \includegraphics[width=\textwidth]{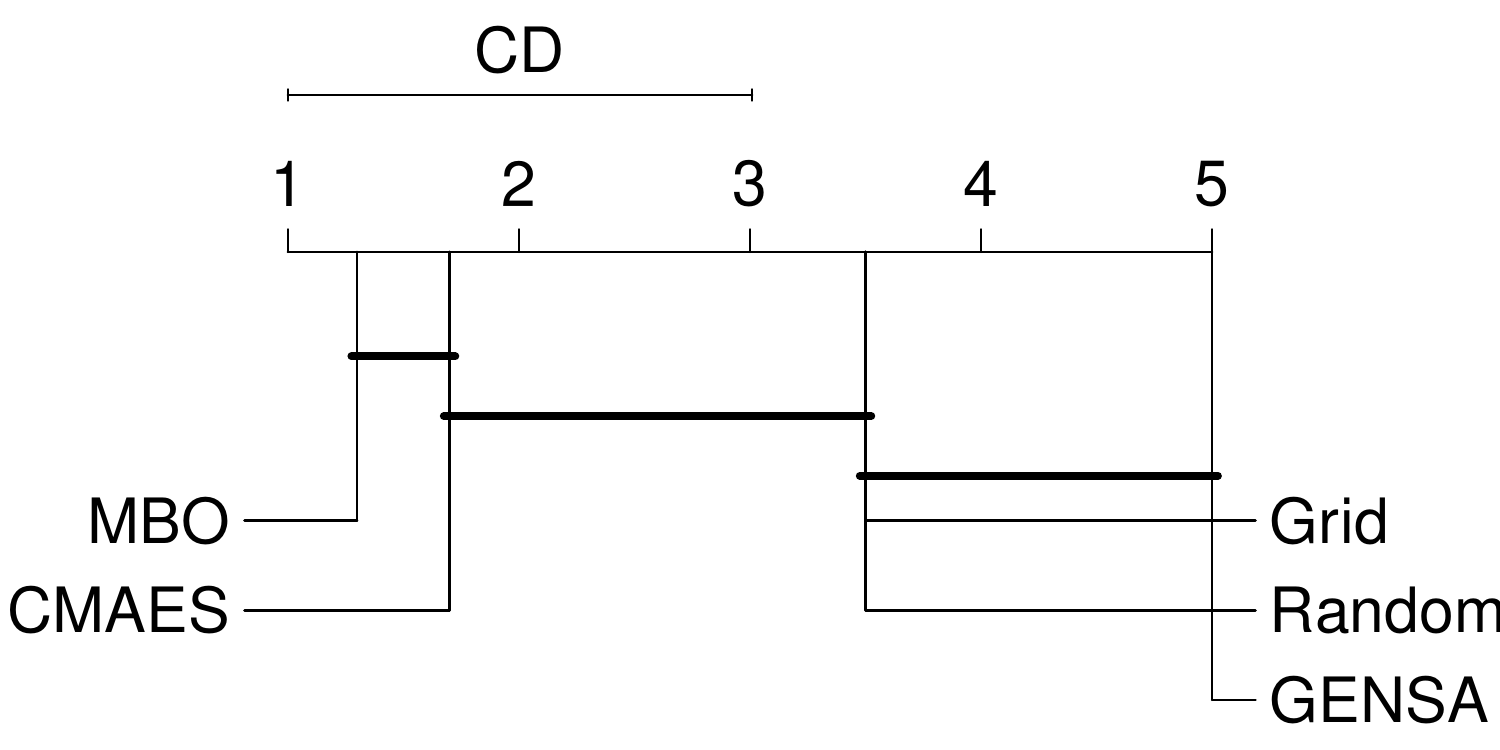}
    \caption{HPO $2D$.}
    \label{fig:hpo_2_cd}
  \end{subfigure}
  \hfill
  \begin{subfigure}[t]{0.32\linewidth}
    \centering
    \includegraphics[width=\textwidth]{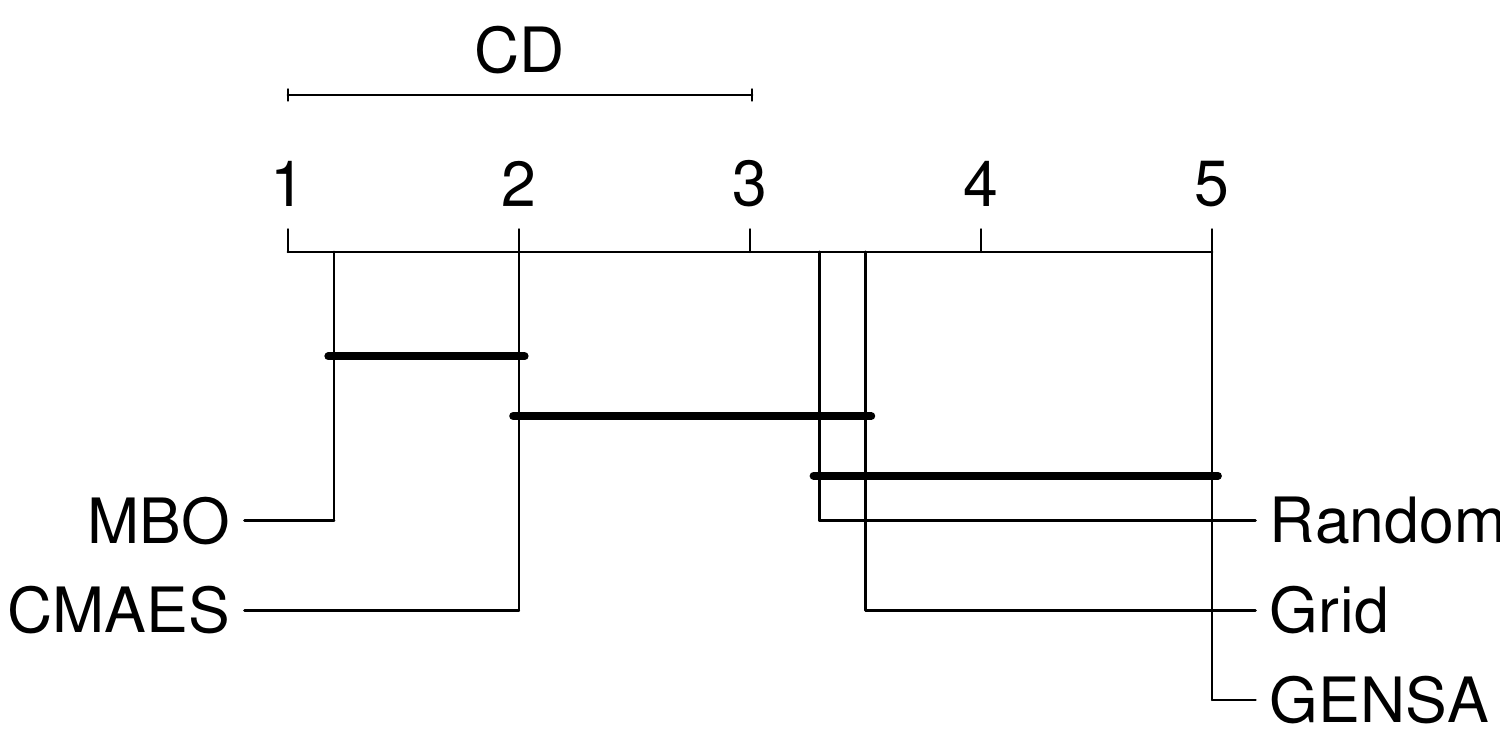}
    \caption{HPO $3D$.}
    \label{fig:hpo_3_cd}
  \end{subfigure}
  \hfill
  \begin{subfigure}[t]{0.32\linewidth}
    \centering
    \includegraphics[width=\textwidth]{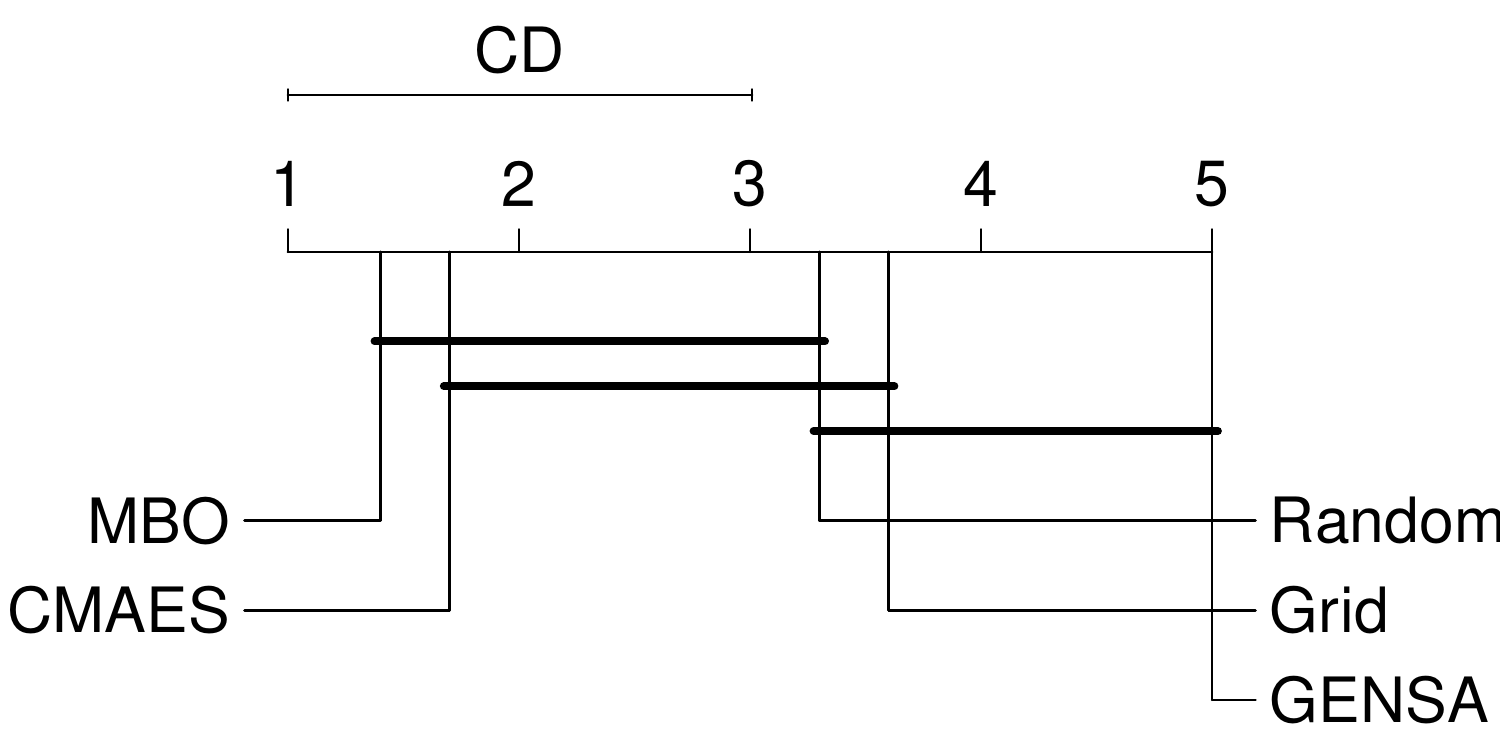}
    \caption{HPO $5D$.}
    \label{fig:hpo_5_cd}
  \end{subfigure}
  \caption{Critical differences plots for mean ranks of optimizers on BBOB and HPO problems split with respect to the dimensionality.}
  \label{fig:bbob_hpo_cd}
\end{figure}

For each BBOB problem, we computed optimizer rankings based on the average final performance (best target value of an optimizer run averaged over replications).
\Cref{fig:bbob_2_cd,fig:bbob_3_cd,fig:bbob_5_cd} visualize the differences in rankings on the BBOB problems split for the dimensionality.
Friedman tests indicated overall significant differences in rankings ($2D:$ $\chi^2(4) = 154.55, p < 0.001$, $3D:$ $\chi^2(4) = 219.16, p < 0.001$, $5D:$ $\chi^2(4) = 258.69, p < 0.001$).
We observe that \texttt{MBO} and \texttt{CMAES} perform well throughout all three dimensionalities, whereas \texttt{GENSA} only is significantly better than \texttt{Grid} or \texttt{Random} for dimensionalities $3$ and $5$.
Moreover, \texttt{Grid} only falls behind \texttt{Random} for the $5D$ problems.

\Cref{fig:hpo_2_cd,fig:hpo_3_cd,fig:hpo_5_cd} analogously visualize differences in rankings on the HPO problems split for the dimensionality.
Friedman tests indicated overall significant differences in rankings ($2D:$ $\chi^2(4) = 36.32, p < 0.001$, $3D:$ $\chi^2(4) = 34.32, p < 0.001$, $5D:$ $\chi^2(4) = 34.80, p < 0.001$).
Again, \texttt{MBO} and \texttt{CMAES} perform well throughout all three dimensionalities.
Notably, \texttt{GENSA} shows lacklustre performance regardless of the dimensionality, failing to outperform \texttt{Grid} or \texttt{Random}.
Similarly as on the BBOB problems, \texttt{Grid} tends to fall behind \texttt{Random} for the higher-dimensional problems.
We do want to note that critical difference plots for the HPO problems are somewhat underpowered when compared to the BBOB problems due to the difference in the number of benchmark problem which results in larger critical distances, as seen in the figures.

\begin{figure}[!t]
  \centering
  \begin{subfigure}[t]{\linewidth}
    \centering
    \includegraphics[width=\textwidth]{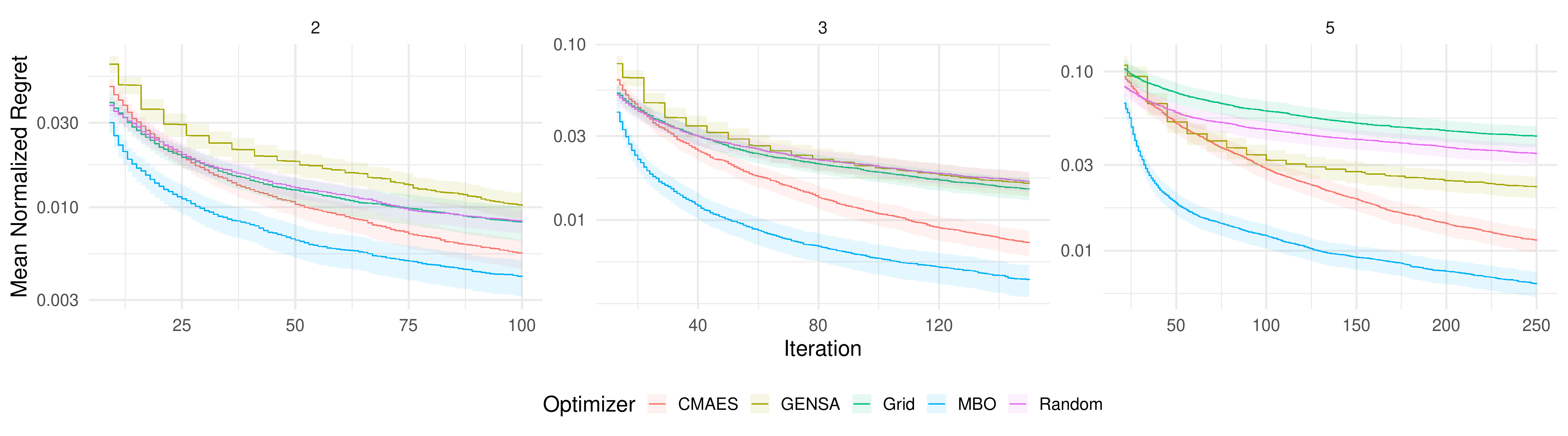}
    \caption{BBOB.}
    \label{fig:bbob_agg_normalized_regret}
  \end{subfigure}
  \vfill
  \begin{subfigure}[t]{\linewidth}
    \centering
    \includegraphics[width=\textwidth]{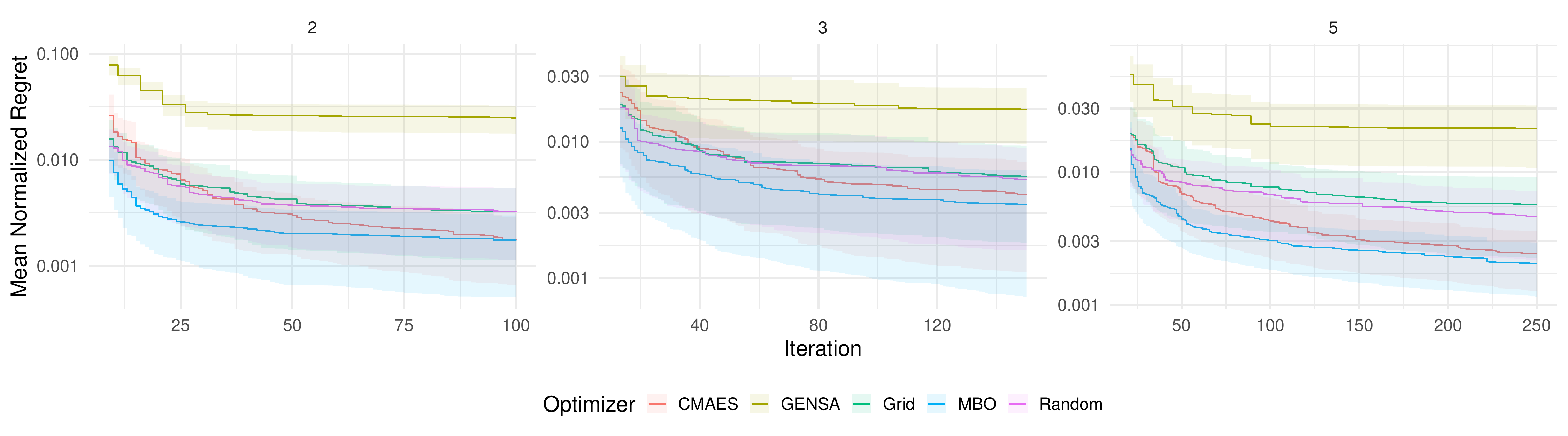}
    \caption{HPO.}
    \label{fig:hpo_agg_normalized_regret}
  \end{subfigure}
  \caption{Anytime mean normalized regret of optimizers on BBOB and HPO problems averaged over replications split for the dimensionality of problems. Ribbons represent standard errors. The x-axis starts after $8\%$ of the optimization budget has been used (initial \texttt{MBO} design).}
  \label{fig:bbob_hpo_agg_normalized_regret}
\end{figure}

In \Cref{fig:bbob_hpo_agg_normalized_regret}, we visualize the anytime performance of optimizers by the mean normalized regret averaged over replications split for the dimensionality of problems.
The normalized regret is defined for an optimizer trace on a benchmark problem as the distance of the current best solution to the overall best solution found across all optimizers and replications, scaled by the overall range of empirical solution values for this benchmark problem.
We choose this metric due to the theoretical optimal solutions being unknown for HPO problems, and apply it to both BBOB and HPO problems to enable performance comparisons.
We observe strong anytime performance of \texttt{MBO} and \texttt{CMAES} on both BBOB and HPO problems regardless their dimensionality.
\texttt{GENSA} shows good performance on the $5D$ BBOB problems but shows poor anytime performance on HPO problems in general.
Differences in anytime performance are less pronounced on the HPO problems, although we do want to note that the width of the standard error ribbons is strongly influenced by the number of benchmark problems.

\begin{figure}[!t]
  \centering
  \begin{subfigure}[t]{0.49\linewidth}
    \centering
    \includegraphics[width=0.75\textwidth]{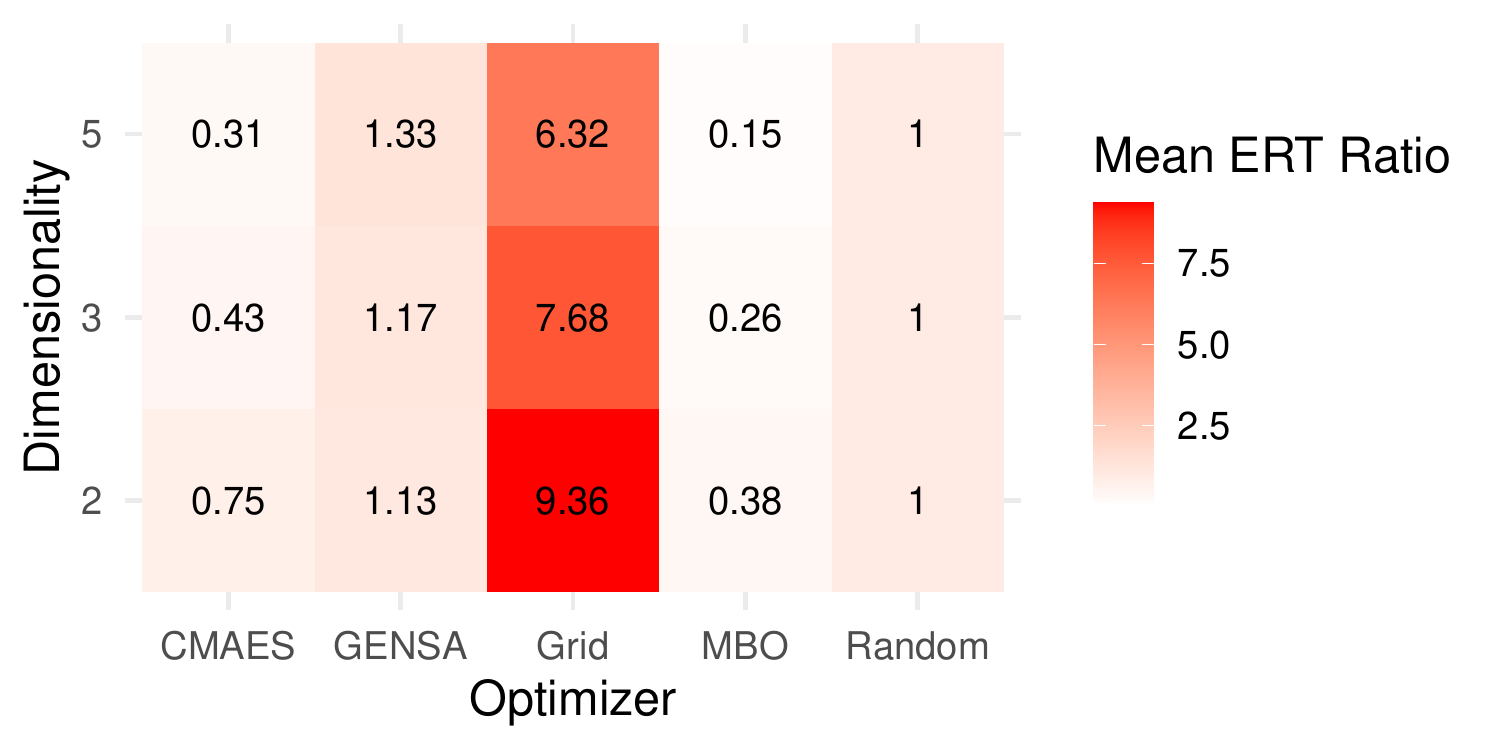}
    \caption{BBOB.}
    \label{fig:bbob_erts_rs_50_agg}
  \end{subfigure}
  \hfill
  \begin{subfigure}[t]{0.49\linewidth}
    \centering
    \includegraphics[width=0.75\textwidth]{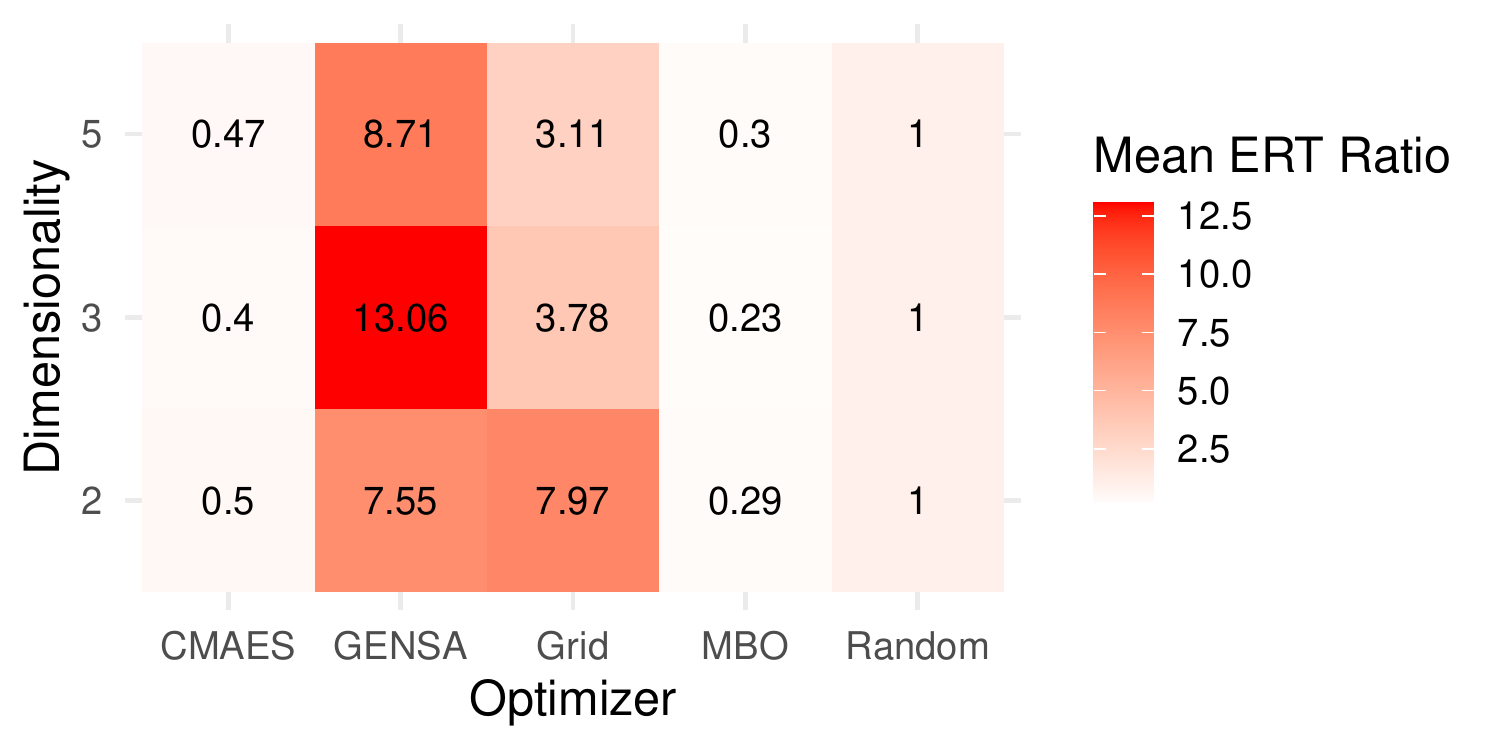}
    \caption{HPO.}
    \label{fig:hpo_erts_rs_50_agg}
  \end{subfigure}
  \caption{Average ERT ratios (optimizers to \texttt{Random}) for HPO and BBOB problems.}
  \label{fig:erts}
\end{figure}

As an additional performance evaluation, we calculated the Expected Running Time (ERT) \cite{hansen2010ert}. In essence, for a given algorithm and problem, the ERT is defined as $\text{ERT} = \frac{1}{n}\sum_{i = 1}^{10}{\text{FE}_i}$, where $n$ is the number of repetitions which are able to reach a specific target, $i$ refers to an individual repetition, and $\text{FE}_i$ denotes the number of function evaluations used.
We investigated the ERT of optimizers with the target given as the median of the best \texttt{Random} solutions (using $50D$ evaluations) over the ten replications per benchmark problem.
We choose this (for BBOB unusual) target due to (1) the theoretical optimum of HPO problems being unknown and (2) \texttt{Random} being considered a strong baseline in HPO \cite{bergstra_algorithms_2011}.
To bring all ERTs on the same scale, we computed the ERT ratios between optimizers and \texttt{Random} per benchmark problem which further allows us to aggregate these ratios over benchmark problems\footnote{Following \cite{kerschke2019automated}, optimizers that did not meet the target in any run were assigned an ERT of the worst ERT on a benchmark problem multiplied by a factor of $10$.}.
We visualize these aggregated ERT ratios separately for the dimensionality of benchmark problems in \Cref{fig:erts}.
We observe that average ERT ratios of \texttt{MBO} and \texttt{CMAES} are comparably similar for BBOB and HPO problems although the tendency that these optimizers become even more efficient with increasing dimensionality is less pronounced on the HPO problems.
\texttt{Grid} generally falls behind and \texttt{GENSA} shows lacklustre performance on HPO.

\section{ELA Feature Space Analysis} \label{sec:ela_analysis}

For each HPO and BBOB problem, we use $50D$ points sampled by LHS (Min-Max) as an initial design for computing ELA features.
We normalize the search space to the unit cube and standardize objective function values per benchmark problem ($(y - \hat{\mu}) / \hat{\sigma}$) prior to calculating ELA features.
This is done to counter potential artefacts that could be seen in ELA features solely due to different value ranges in decision and, in particular, in objective space.
We calculate the feature sets \texttt{ela\_meta}, \texttt{ic}, \texttt{ela\_distr}, \texttt{nbc} and \texttt{disp}, which were introduced in \Cref{sec:background}, using the \texttt{flacco} R package \cite{kerschke2019comprehensive}.

\begin{figure}[!t]
  \centering
  \begin{subfigure}[t]{0.345\linewidth}
    \centering
    \includegraphics[width=\textwidth]{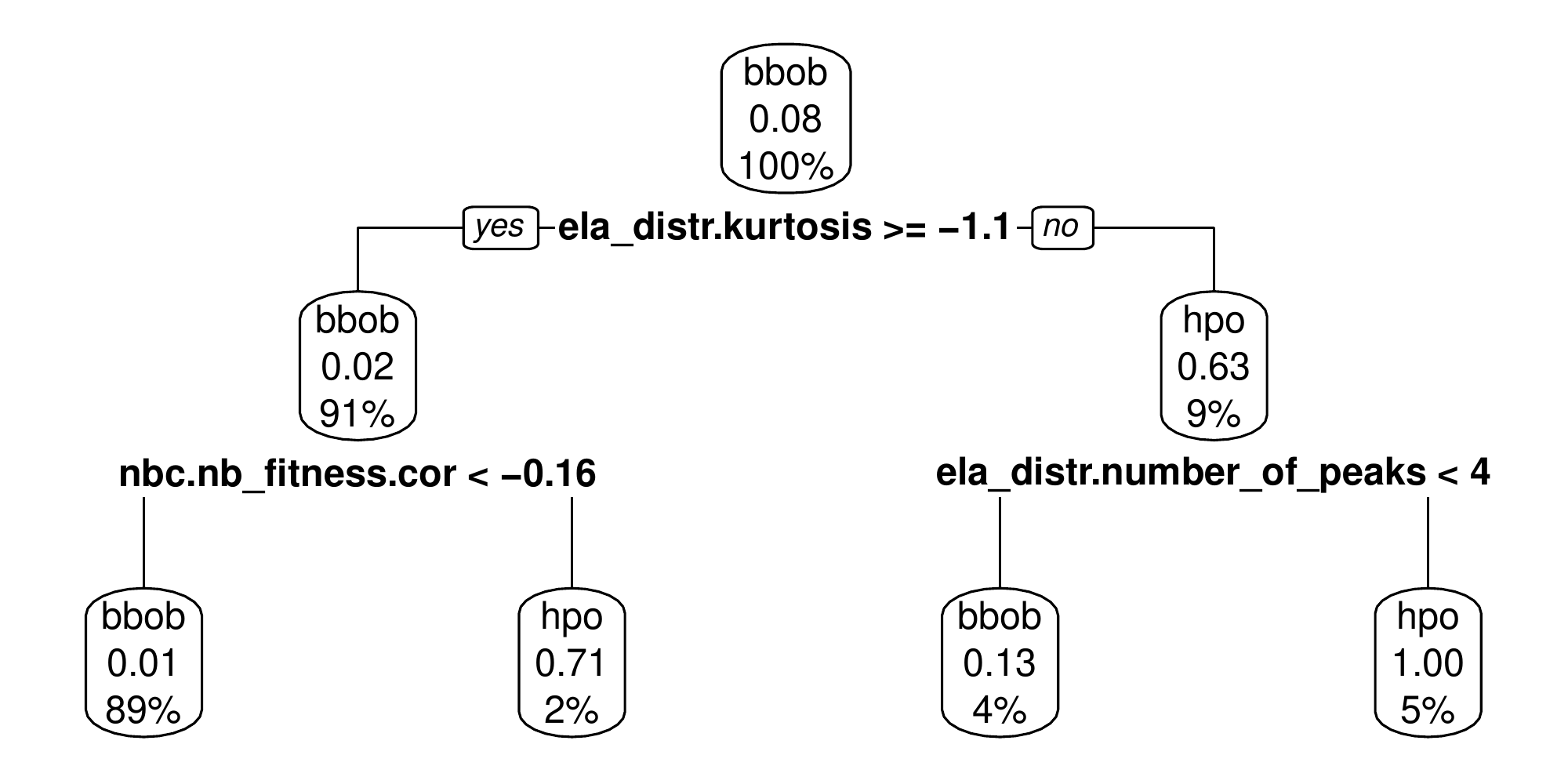}
    \caption{HPO vs. BBOB.}
    \label{fig:classify_type_rp}
  \end{subfigure}
  \hfill
  \begin{subfigure}[t]{0.645\linewidth}
    \centering
    \includegraphics[width=\textwidth]{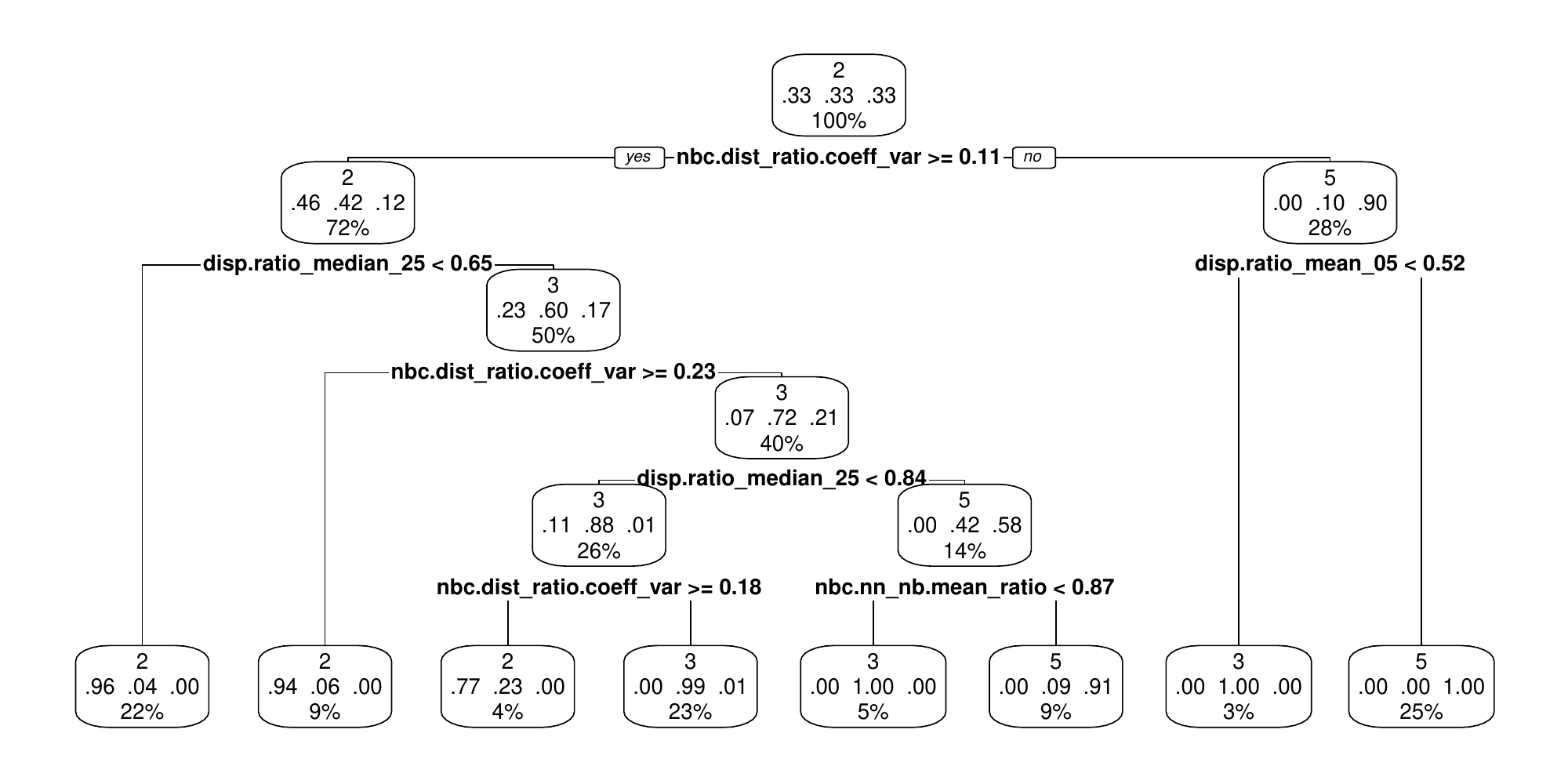}
    \caption{Dimensionality.}
   \label{fig:classify_dim_bbob_rp}
   \end{subfigure}
  \caption{
  Decision trees for classifying benchmark problems into HPO or BBOB problems (left) and classifying the dimensionality of BBOB problems (right).
  }
  \label{fig:classify_trees}
\end{figure}

To answer the question whether ELA can be used to distinguish HPO from BBOB problems, we construct a binary classification task using ELA features to predict the label ``HPO'' vs. ``BBOB''.
We use a decision tree and estimate the generalization error via 10 times repeated 10-fold cross-validation (stratified for the target).
We obtain an estimated classification error of $3.54\%$.
\Cref{fig:classify_type_rp} illustrates the decision tree obtained after training on all data.
We observe that only few ELA features are needed to correctly classify problems:
HPO problems tend to exhibit a lower \texttt{ela\_distr.kurtosis} combined with more \texttt{ela\_distr.number\_of\_peaks} or show a higher \texttt{nbc.nb\_fitness.cor} than BBOB problems if the first split with respect to the kurtosis has not been affirmed.
This finding is supported by visualizations of the $2D$ HPO problems, which we present in our \online, i.e., most $2D$ HPO problems have large plateaus resulting in negative kurtosis.

To answer the question whether dimensionality is a different concept for HPO compared to BBOB problems\footnote{For HPO problems, it is a priori often unclear whether a change in a parameter value also results in relevant objective function changes, i.e., the intrinsic dimensionality of a HPO problem may be lower than the number of hyperparameter suggests.} we perform the following analysis:
We construct a classification task using ELA features to predict the dimensionality of the problem but only use the BBOB subset for the training of a decision tree.
We estimate the generalization error via 10 times repeated 10-fold cross-validation (stratified for the target) and obtain an estimated classification error of $7.39\%$.
We then train the decision tree on all BBOB problems (illustrated in \Cref{fig:classify_dim_bbob_rp}) and determine the holdout performance on the HPO problems and obtain a classification error of $10\%$.
Only few ELA features of the \texttt{disp} and \texttt{nbc} group are needed to predict the dimensionality of problems with high accuracy.
Intuitively, this is sensible, due to \texttt{nbc} features involving the calculation of distance metrics (which themselves should be affected by the dimensionality) and both \texttt{nbc} and \texttt{disp} features being sensible to the multimodality of problems \cite{kerschke2015detecting,lunacek2006dispersion} which should also be affected by the dimensionality.
Based on the reasonable good hold-out performance of the classifier on the HPO problems, we conclude that ``dimensionality'' is a similar concept for BBOB and HPO problems.

\begin{figure}[!t]
  \centering
  \includegraphics[width=\textwidth]{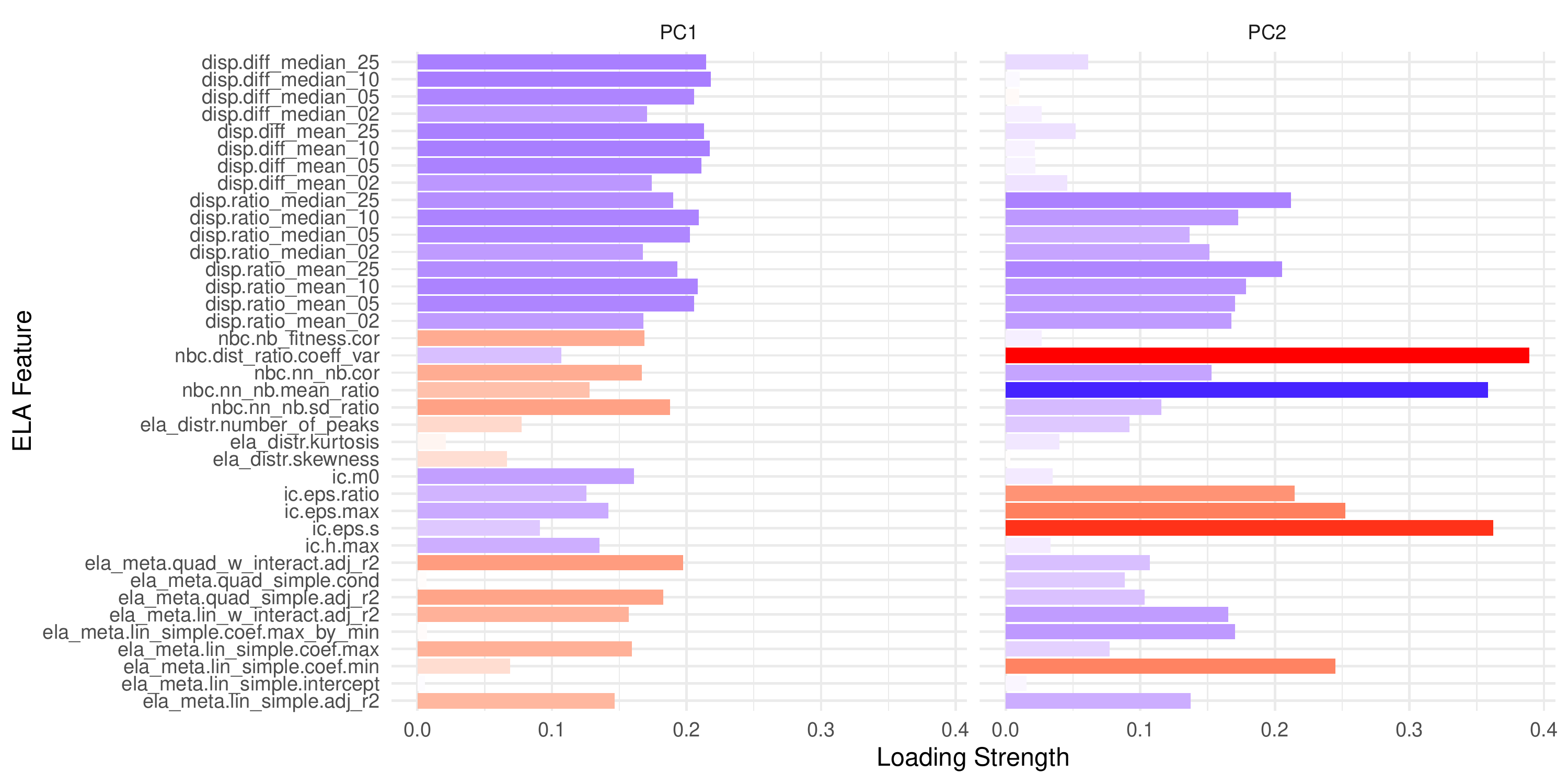}
  \caption{Factor loadings of ELA features on the first two principle components. Blue indicates a positive loading, whereas red indicates a negative loading.}
  \label{fig:ela_pca_loading}
\end{figure}

To gain insight on a meta-level, we performed a PCA on the scaled and centered ELA features of both the HPO and BBOB problems.
To ease further interpretation, we select a two component solution that explains roughly $60\%$ of the variance.
\Cref{fig:ela_pca_loading} summarizes factor loadings of ELA features on the first two principle components.
Most \texttt{disp} features show a medium positive loading on PC1, whereas some \texttt{nbc} show medium negative loadings.
\texttt{ela\_meta} features, including $R^2$ measures of linear and quadratic models, also exhibit medium negative loadings on PC1.
We therefore summarize PC1 as a latent dimension that mostly reflects multimodality of problems.
Regarding PC2, three features stand out with strong loadings: \texttt{nbc.dist\_ratio.coeff\_var}, \texttt{nbc.nn\_nb.mean\_ratio} and \texttt{ic.eps.s}.
Moreover, \texttt{disp.ratio\_*} features generally have a medium negative loading.
We observe that all features used by the decision tree in \Cref{fig:classify_dim_bbob_rp} also have comparably large loadings on PC2.
Therefore, we summarize PC2 as an indicator of the dimensionality of problems.

\begin{figure}[!t]
  \centering
  \includegraphics[width=\textwidth]{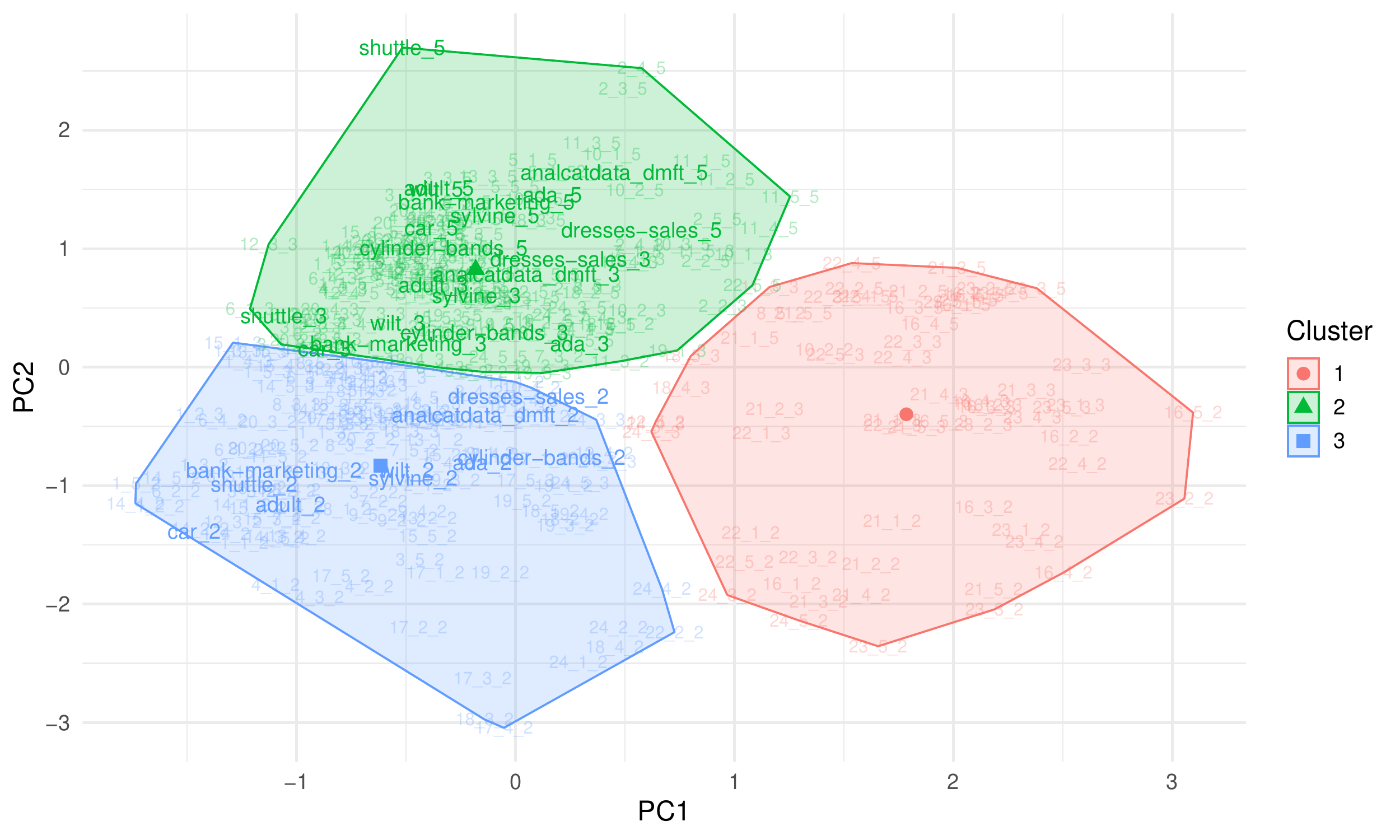}
  \caption{Cluster analysis of BBOB and HPO problems on the first two principle component scores in ELA feature space.}
  \label{fig:ela_cluster}
\end{figure}

We then performed k-means clustering on the two scaled and centered principal component scores.
A silhouette analysis suggested the selection of three clusters.
In \Cref{fig:ela_cluster}, we visualize the assignment of HPO and BBOB problems to these three clusters.
Labels represent IDs of BBOB  and HPO problems.
We observe that the dimensionality of problems is almost perfectly reflected in the PC2 alignment.
Cluster 2 and 3 can be mostly distinguished along PC2 (cluster 3 contains low dimensional problems and cluster 2 contains higher dimensional problems) whereas cluster 1 contains problems with large PC1 values.
HPO problems are exclusively assigned to cluster 2 or 3, exhibiting low variance with respect to their PC1 score, with the PC1 values indicating low multimodality.

\begin{figure}[!t]
  \centering
  \begin{subfigure}[t]{0.32\linewidth}
    \centering
    \includegraphics[width=\textwidth]{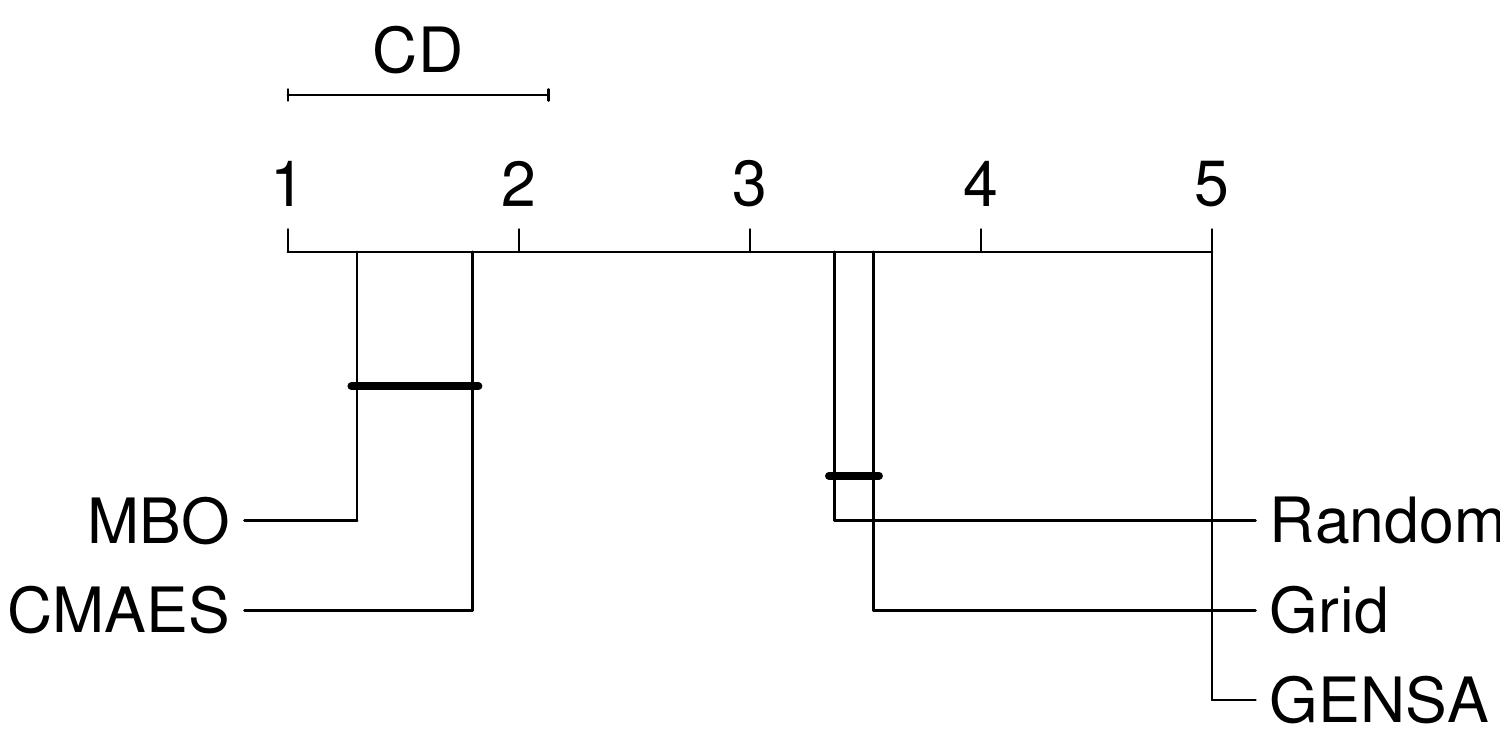}
    \caption{HPO.}
    \label{fig:hpo_all_cd}
  \end{subfigure}
  \hspace{0.1\linewidth}
  \begin{subfigure}[t]{0.32\linewidth}
    \centering
    \includegraphics[width=\textwidth]{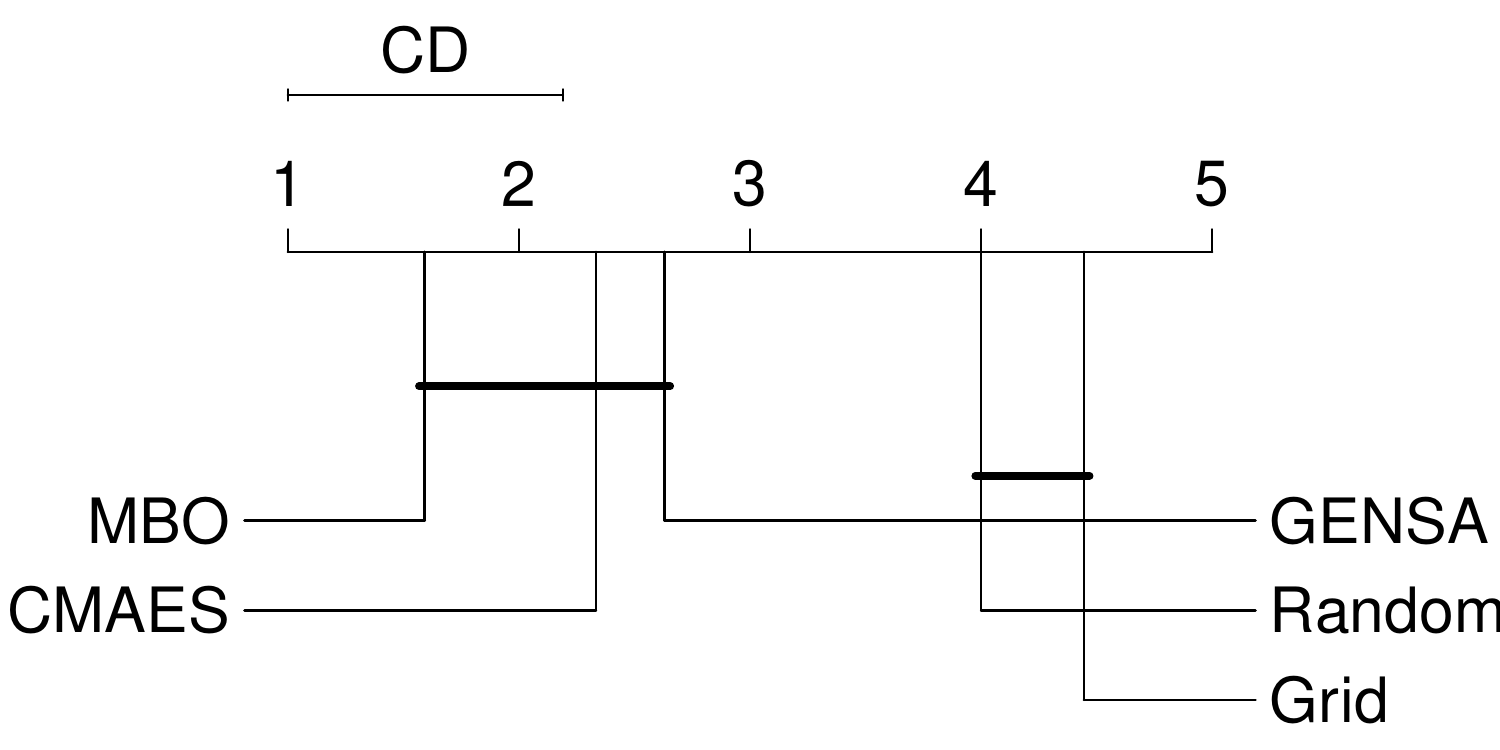}
    \caption{Nearest BBOB.}
    \label{fig:bbob_nearest_hpo_cd}
  \end{subfigure}
  \caption{Critical differences plots for mean ranks of optimizers on all HPO problems (left) and the subset of nearest BBOB problems.}
  \label{fig:hpo_bbob_nearest_cd}
\end{figure}

As a final analysis we determined the nearest BBOB neighbors of the HPO problems (in ELA feature space based on the cluster analysis, i.e., minimizing the Euclidean distance over the first two principal component scores).
For a complete list, see our \online.
We again computed optimizer rankings based on the average final performance of the optimizers (over the replications), but this time for all HPO problems (regardless their dimensionality) and the subset of BBOB problems that are closest to the HPO problems in ELA feature space (see \Cref{fig:hpo_bbob_nearest_cd}).
Friedman tests indicated overall significant differences in rankings for both HPO ($\chi^2(4) = 104.99, p < 0.001$) and nearest BBOB ($\chi^2(4) = 61.01, p < 0.001$) problems.
We observe similar optimizer rankings, with \texttt{MBO} and \texttt{CMAES} outperforming \texttt{Random} or \texttt{Grid}, indicating that closeness in ELA feature space somewhat translates to optimizer performance.
Nevertheless, we do have to note that \texttt{GENSA} exhibits poor performance on the HPO problems compared to the nearest BBOB problems.
We hypothesize that this may be caused by the performance of \texttt{GENSA} being strongly influenced by its hyperparameter configuration itself and provide an initial investigation in our \online.

\section{Conclusion} \label{sec:conclusion}

In this paper, we characterized the landscapes of continuous hyperparameter optimization problems using ELA.
We have shown that ELA features can be used to (1) accurately distinguish HPO from BBOB problems and (2) classify the dimensionality of problems.
By performing a cluster analysis in ELA feature space, we have shown that our HPO problems mostly position themselves with BBOB problems of little multimodality, mirroring the results of \cite{pushak2018algorithm,pushak2022automl}.
Determining the nearest BBOB neighbor of HPO problems in ELA feature space allowed us to investigate performance differences of optimizers with respect to HPO problems and their nearest BBOB problems and we observed comparably similar performance.
We believe that this work is an important first step in identifying BBOB problems that can be used in lieu of real HPO problems when, for example, configuring or developing novel HPO methods.

Our work still has several limitations. 
A major one is that traditional ELA is only applicable to continuous HPO problems, which constitute a minority of real-world problems.
In many practical applications, search spaces include categorical and conditionally active hyperparameters -- so-called hierarchical, mixed search spaces \cite{thornton2013auto}.
In such scenarios, measures such as the number of local optima, fitness-distance correlation or auto-correlation of fitness along a path of a random walk \cite{hains2011revisiting,hernando2013evaluation} can be used to gain insight into the fitness landscape.
Another limitation is that our studied HPO problems all stem from tuning XGBoost, with little variety of comparably low dimensional search spaces, which limits the generalizability of our results.

In future work, we would like to extend our experiments to cover a broader range of HPO settings, in particular different learners and search spaces, but also data sets.
We also want to reiterate that HPO is generally noisy and expensive.
In our benchmark experiments, costly 10-fold cross-validation with a fixed instantiating per data set was employed to reduce noise to a minimal level.
Future work should explore the effect of the variance of the estimated generalization error on the calculation and usage of ELA features which poses a serious challenge for ELA applied to HPO in practice.
Besides, we used logloss as a performance metric which by definition is rather ``smooth'' compared to other metrics such as the classification accuracy (but the concrete choice of performance metric typically depends on the concrete application at hand).
Moreover, ELA requires the evaluation of an initial design, which is very costly in the context of HPO.
In general, HPO often can be performed with evaluations on multiple fidelity levels, i.e., by reducing the size of training data, and plenty of HPO methods make use of this resulting in significant speed-up \cite{li17}.
Future work could explore the possibility of using low fidelity evaluations for the initial design required by ELA and how multiple fidelity levels of HPO affect ELA features.

We consider our work as pioneer work and hope to ignite the research interest in studying the landscape properties of HPO problems going beyond fitness measures.
We envision that, by improved understanding of HPO landscapes and identifying relevant landscape properties, better optimizers may be designed, and eventually instance-specific algorithm selection and configuration for HPO may be enabled.\\

\noindent\textbf{Acknowledgements}
This work was supported by the German Federal Ministry of Education and Research (BMBF) under Grant No. 01IS18036A. H. Trautmann, R. P. Prager and P. Kerschke acknowledge support by the European Research Center for Information Systems (ERCIS). Further, L. Schäpermeier and P. Kerschke acknowledge support by the Center for Scalable Data Analytics and Artificial Intelligence (ScaDS.AI) Dresden/Leipzig.
L. Schneider is supported by the Bavarian Ministry of Economic Affairs, Regional Development and Energy through the Center for Analytics – Data – Applications (ADACenter) within the framework of BAYERN DIGITAL II (20-3410-2-9-8).

\bibliographystyle{splncs04}
\bibliography{refs}

\end{document}